\documentclass[letterpaper, 10 pt, journal, twoside]{ieeetran}

\IEEEoverridecommandlockouts                              

\usepackage{graphicx}
\usepackage{mathptmx} 
\usepackage{times} 
\usepackage{amsmath} 
\usepackage{amssymb}  
\usepackage{wasysym}
\usepackage{subfigure}
\usepackage{hyperref}

\DeclareMathAlphabet{\mathcal}{OMS}{cmsy}{m}{n}
 
\title{SKID RAW: Skill Discovery from Raw Trajectories}

\author{Daniel Tanneberg$^{1}$ and Kai Ploeger$^{1}$ and Elmar Rueckert$^{2,1}$ and Jan Peters$^{1,3}$
\thanks{Manuscript received: December 2, 2020; Revised February 11, 2021; Accepted March 8, 2021.}
\thanks{This paper was recommended for publication by Dana Kulic upon evaluation of the Associate Editor and Reviewers' comments.}
\thanks{This project has received funding from the European Union's Horizon 2020 research and innovation programme under grant agreement No. \#713010 (GOAL-Robots) and No. \#640554 (SKILLS4ROBOTS), and from the Deutsche Forschungsgemeinschaft (DFG, German Research Foundation) - No \#430054590 (TRAIN).
This research was supported by NVIDIA.}
\thanks{$^{1}$Intelligent Autonomous Systems, Technische Universit\"at Darmstadt,}%
\thanks{$^{2}$Institute for Robotics and Cognitive Systems, Universit\"at zu L\"ubeck}%
\thanks{$^{3}$Robot Learning Group, Max-Planck Institute for Intelligent Systems}%
\thanks{Digital Object Identifier (DOI): see top of this page.}
}

\begin{document}

\maketitle

\begin{abstract}
Integrating robots in complex everyday environments requires a multitude of problems to be solved.
One crucial feature among those is to equip robots with a mechanism for teaching them a new task in an easy and natural way.
When teaching tasks that involve sequences of different skills, with varying order and number of these skills, it is desirable to only demonstrate full task executions instead of all individual skills.
For this purpose, we propose a novel approach that simultaneously learns to segment trajectories into reoccurring patterns and the skills to reconstruct these patterns from unlabelled demonstrations without further supervision.
Moreover, the approach learns a skill conditioning that can be used to understand possible sequences of skills, a practical mechanism to be used in, for example, human-robot-interactions for a more intelligent and adaptive robot behaviour.
The Bayesian and variational inference based approach is evaluated on synthetic and real human demonstrations with varying complexities and dimensionality, showing the successful learning of segmentations and skill libraries from unlabelled data.
\end{abstract}
 

\section{Introduction}
\IEEEPARstart{W}{hile} pattern recognition, the ability to find order and regularities in noisy observations, is a necessary skill for intelligent behaviour, learning to abstract and transfer that information is a crucial ability that goes beyond pattern recognition~\cite{lake2017building}.
Such cognition abilities are also helpful to employ intelligent robots into our everyday life, by lowering the barrier to program the desired robotic behaviour.
While often trained experts can program robots with complex behaviour, this process requires a lot of expert knowledge in different domains, is cost intensive and often limited to special tasks or domains.
Instead, it is easier to teach the robot the desired behaviour instead of programming it and is the main motivation behind the learning from demonstrations paradigm~\cite{argall2009survey}.
Teaching the robot by showing the desired behaviour is not only a more natural and intuitive way of programming the robot, but it also drastically reduces the required expert knowledge of the system and time.

Typically demonstrations consist of single tasks, and one skill, or policy, per such task is learned from these demonstrations.
For more complex tasks, that typically consist of multiple subtasks, the task is often either broken down into demonstrated subtasks or the demonstrations are segmented and labelled afterwards.
Both methods again require specific knowledge, are time consuming, and scale poorly with the number of (sub)tasks and their arrangements.
Subtasks can appear within a demonstration as well as across demonstrations and it can be challenging to decide where to cut or which segments can be considered the same subtask.

A more natural approach is to just demonstrate the full tasks, and let the system automatically learn to segment the full demonstration into subtasks or skills~\cite{lioutikov2017learning,niekum2012learning}.
Such decompositions of movements into different phases were also detected in the primary motor cortex of monkeys performing reaching tasks~\cite{kadmon2019movement}.
When equipping robots with such an automatic segmentation technique, teaching tasks which consist of a sequence of skills becomes easier for non-experts. 
Furthermore, by learning the segmentation, the robot can \textit{understand} the demonstrated tasks by, for example, learning important (sub)goals or which skills are likely to follow each other.
This approach can therefore also be used in human-robot interaction~\cite{goodrich2008human} by, for example, learning the sequence of movements of a human, and using this knowledge to predict the humans future behaviour, which allows for a more intelligent and adaptive robot behaviour.

\begin{figure}
\centering
\includegraphics[width=0.31\textwidth]{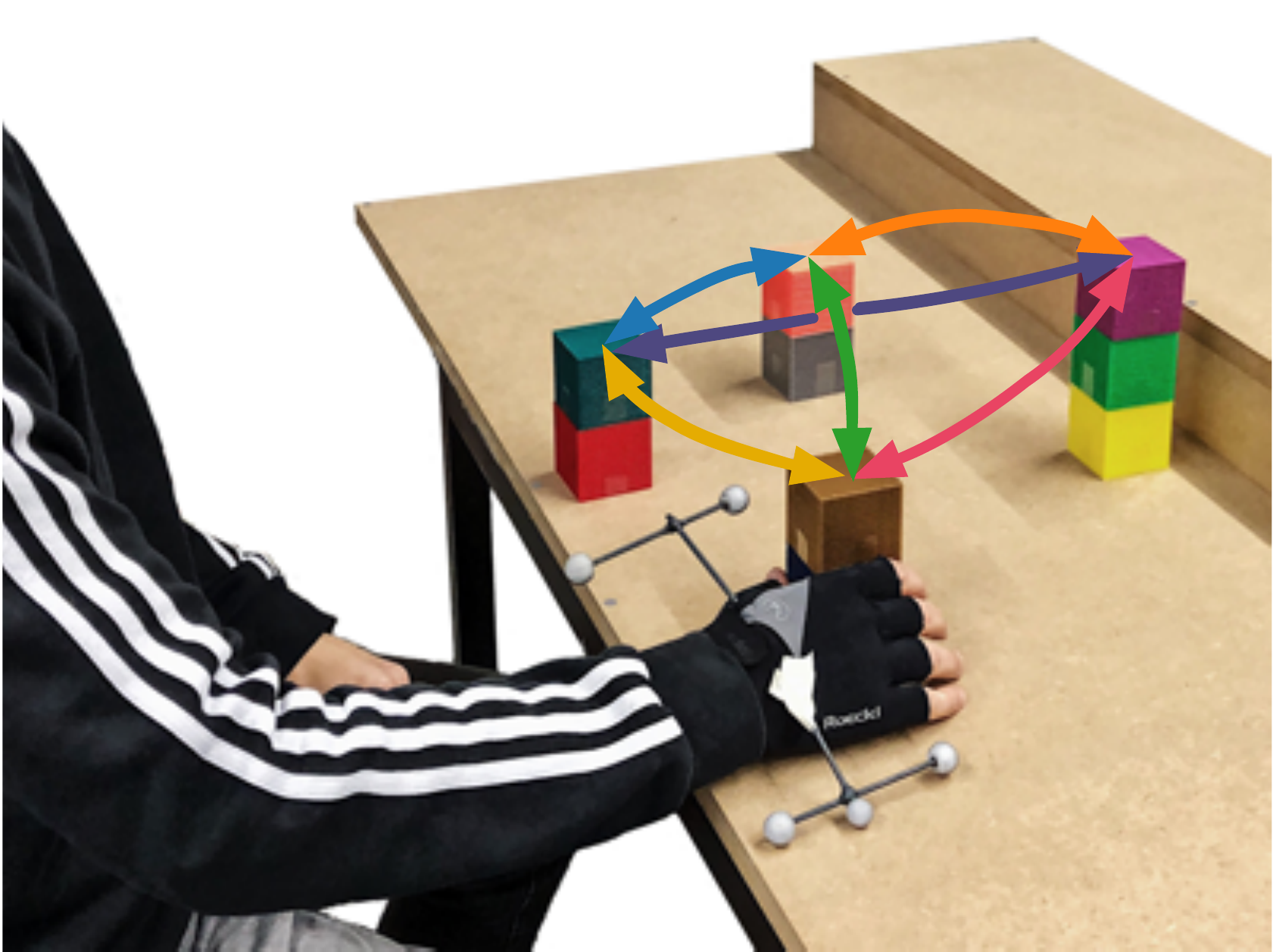}
\caption{Snapshot from the data collection process from the human teaching experiment. 
The human was instructed to move $1$-$3$ cubes from one position to another, while the hand position was tracked.
These unlabelled hand position trajectories consisting of a variable number of moved cubes are used to learn to segment these into skills and the skills simultaneously. 
}
\label{fig:data_record}
\vspace{-12pt}
\end{figure}

\begin{figure*}[t]
\centering
\hfill
\subfigure[Graphical model.]{\includegraphics[width=.20\textwidth]{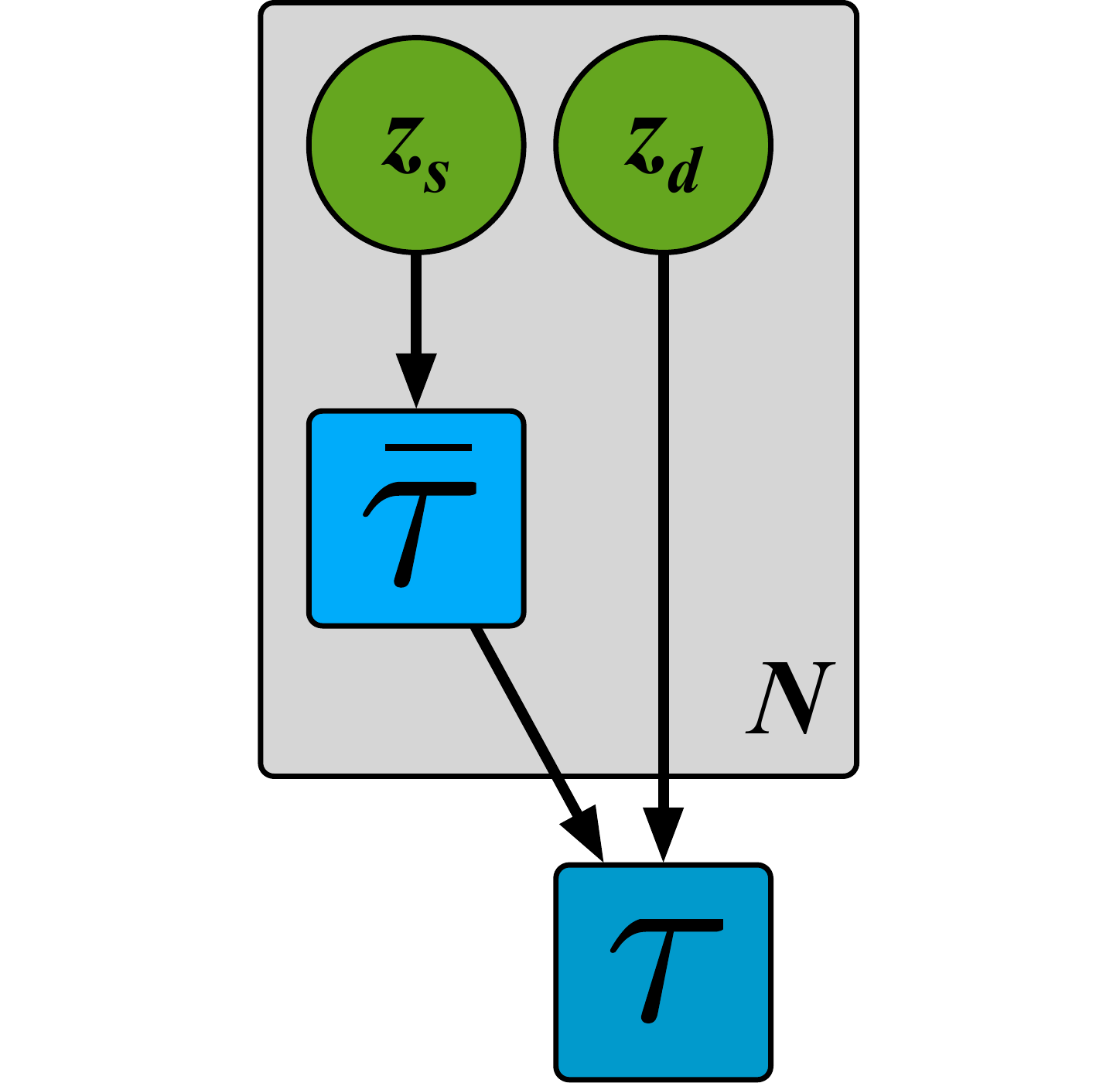}} 
\hspace{50pt}
\subfigure[SKID framework.]{\includegraphics[width=.40\textwidth]{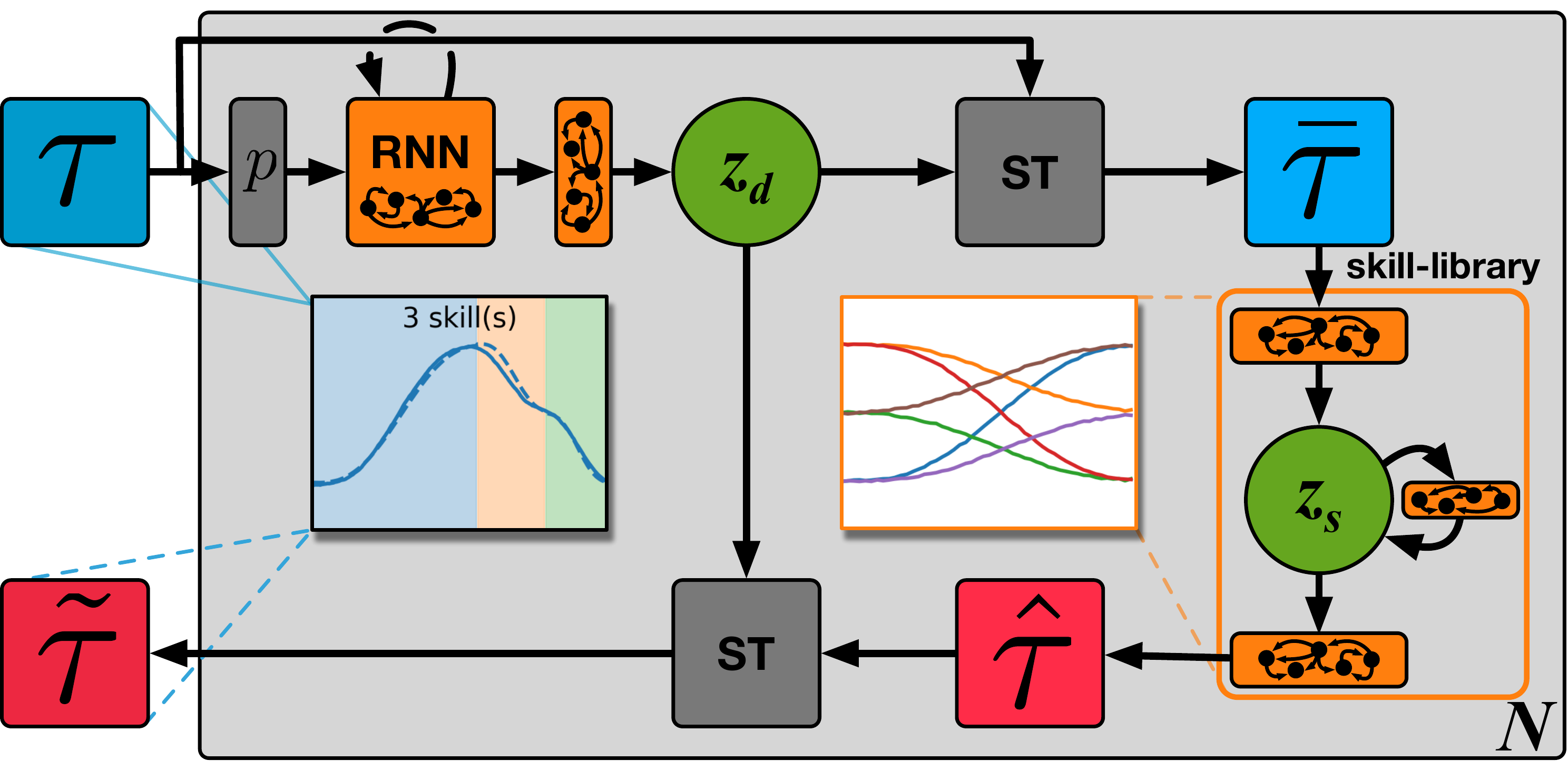}}
\hfill
\vspace{-2pt}
\caption{(a) Graphical model representation of the relationship between the $N$-dimensional latent trajectory description $z = \{z_d , z_s\}$, the sub-trajectory $\bar{\tau}$ and the full trajectory $\tau$.
(b) Sketch of the SKID implementation with the learned neural networks in orange. 
The recurrent neural network (RNN) and the output layer learn the distribution of $z_d$ given the, potentially $p$ transformed, trajectory $\tau$.
Then $z_d$ and the raw trajectory are used in a spatial transformer (ST) to extract the sub-trajectory $\bar{\tau}$.
This sub-trajectory is used to learn the skill library and the approximated sub-trajectory $\hat{\tau}$ is added to the reconstructed trajectory $\tilde{\tau}$ with a spatial transformer.
This process is repeated $N$ times, iteratively segmenting the full trajectory into sub-trajectories utilizing the simultaneously learned skill library.
The two inlays show examples from the 1D synthetic task, where the left inlay shows the original trajectory $\tau$ in blue and the reconstructed trajectory $\tilde{\tau}$ in red.
The coloured background indicates the learned skill $z_s$ used to approximate this segment of size $z_d$.
The right inlay shows the learned skill library.
}
\label{fig:model}
\vspace{-12pt}
\end{figure*}

\subsection{Contribution}
Here, we propose a novel approach for learning simultaneously to segment trajectories into reoccurring patterns and the skills to reproduce them from unlabelled trajectories without supervision.
Our approach is based on the variational autoencoder (VAE)~\cite{kingma2014auto} framework and the iterative concept of AIR~\cite{eslami2016attend} to explain sub-parts of the given data per iteration.
A similar iterative approach is investigated in the CompILE framework~\cite{kipf2019compile}, but without an interpretable latent space, skill conditioning and length-independent skills.
In contrast to related approaches~\cite{osa2020goal,noseworthy2020task} that learn skills with VAEs from single-skill trajectories, our approach operates on trajectories with a varying and unknown number of skills per demonstration and learns to segment the trajectory and the skill library simultaneously.

Furthermore, the proposed method requires less initial knowledge like heuristic cutting points~\cite{lioutikov2017learning}, expert domain knowledge~\cite{steinmetz2019intuitive}, segmented and labelled  demonstrations~\cite{manschitz2015learning}, pretraining~\cite{shankar2020learning}, and segmentation and skill library are trained simultaneously instead of subsequently~\cite{niekum2012learning,su2018learning,zhang2018complex} to use the skill knowledge for segmentation.
Trajectory segmentation shares similarities with switching systems and hierarchical skill discovery, investigated, for example, in (inverse) reinforcement learning settings~\cite{ranchod2015nonparametric,kroemer2015towards}, hybrid systems~\cite{burke2020hybrid} and compositional program induction~\cite{burke2019explanation}.
 
We show that our approach can be used to learn skills from unlabelled demonstrations of full task executions involving a varying and unknown number of skills per demonstration, and that the learned model can also be used to understand the relation between skills and, hence, predict possible future skills for adaptive behaviour.

\section{Discovering Skills from Raw Trajectories}
The proposed skill discovery (SKID) approach is inspired by the AIR~\cite{eslami2016attend} model for images.
AIR is a generative model that learns to reconstruct visual scenes by learning the properties of individual objects to render the scene.
We take this approach as inspiration to construct a similar generative model for trajectories, that learns to segment raw trajectories into reoccurring patterns (subtasks) and the individual skills to reconstruct these simultaneously and without supervision.

We assume that a trajectory $\tau$ is composed of $N$ sub-trajectories $\bar{\tau}$ that we call skills. 
Like AIR~\cite{eslami2016attend} did for images, we take a Bayesian perspective of trajectory understanding and treat it as inference in a generative model.
In general, for a given trajectory $\tau$, the model parametrized by $\theta$ is given by $p^{\tau}_{\theta}(\tau | z) p^z_{\theta}(z)$, where the prior $p^z_{\theta}(z)$ over the latents $z$ captures the trajectory assumptions, and the likelihood $p^{\tau}_{\theta}(\tau | z)$ describes how the latent trajectory description is composed into the full trajectory.
We are interested in recovering the trajectory description $z$, given by computing the posterior
\begin{align}
p(z | \tau) = p^{\tau}_{\theta}(\tau | z) p^z_{\theta}(z) / p(\tau) \ .
\end{align}

As we assume a trajectory $\tau$ consists of $N$ sub-trajectories or skills, the trajectory description is structured into a sequence of $z^i$.
Each $z^i$ is a structure that describes the properties of one skill, here, duration and skill type, i.e., $z = \{z_d , z_s\}$.
The duration of a skill $z_d$ is used for the segmentation and the skill type $z_s$ is a discrete identifier for each skill.
Figure~\ref{fig:model}(a) shows this relation between $\tau$ and $z^i$ in a graphical model.
The generative model is given as
\begin{align}
p_{\theta}(\tau) &= \sum_{n=1}^{N} \int p^z_{\theta}(z^n | z^{n-1}) p^{\tau}_{\theta}(\tau | z) dz \ ,
\end{align}
where $N$ is given by the sequence of $z^n_d$ summed up until a given threshold is reached, i.e., $\sum_{n=1}^{N} z^n_d \geq t_{\epsilon}$.
The number of skills ($N$) in a trajectory is dependent on the number of available skills in the library ($S$) and learned without additional feedback from the reconstruction loss.
The condition of $z^n$ on $z^{n-1}$ allows to learn a conditional sequencing of the skills, i.e., learning which skills are likely to follow after each other.
This conditioning is necessary to sample valid trajectories from the generative model, i.e., trajectories without big jumps between the individual skills.
Additionally, by learning this conditioning, the model can be used to \textit{understand} the presented task trajectories, e.g., like predicting which skills are likely to follow each other for adaptive robot behaviour, or be used in planning algorithms when solving new tasks with the learned skills.

\begin{figure*}[t]
\centering
\includegraphics[width=0.90\textwidth]{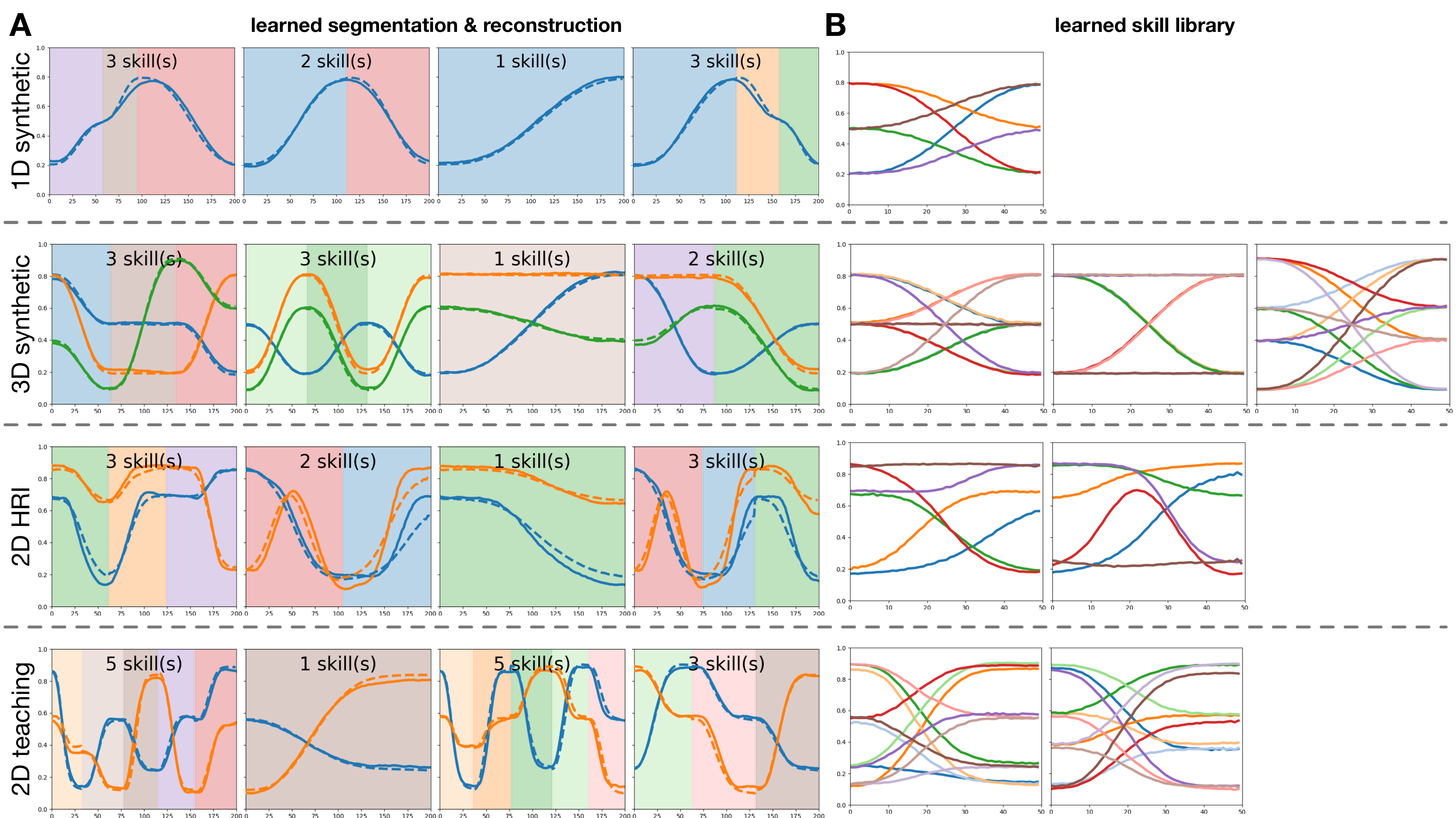}
\caption{Showing the learned segmentations and skill libraries by SKID for the four datasets (see Section~\ref{sec:exps}). 
\textbf{A} shows segmented and reconstructed sample trajectories.
Line colours indicate trajectory dimensions, where solid lines show the original data and dashed lines show the reconstructions. 
The coloured area highlights the segmentation and the colour indicates which skill is used for that segment. 
\textbf{B} shows the learned skill library with one panel per trajectory dimension and one colour for each skill.
}
\label{fig:results_overview}
\vspace{-12pt}
\end{figure*}

\subsection{SKID Instantiation}
We use an amortized variational approximation of the true posterior $p(z | \tau)$ which learns a parametrized distribution $q_{\phi}(z | \tau)$ by minimizing the Kullback–Leibler divergence between $KL(q_{\phi}(z | \tau) ||  p(z | \tau))$.
The inference model $q_{\phi}(z | \tau)$, parametrized by $\phi$ is realized as a recurrent neural network to take previous $z_d$'s into account, i.e., allowing the model to remember which part of the trajectory have been explained already.
Before the trajectory $\tau$ is fed into the recurrent network, it is preprocessed by a function $p$.
In our experiments, we tested with $p$ as the identity function and $p$ outputting the mean velocity of $\tau$ over all dimensions for each timestep.
The mean velocity performed slightly better and was used in all evaluations.
The skill duration $z_d$ is modelled with a Gaussian distribution with a given prior, i.e., $z_d \sim \mathcal{N}(\mu_{d},\sigma^2_{d})$, fed into a \texttt{sigmoid} activation to get the duration as the fraction of the trajectory length.

To extract the part of the trajectory $\tau$ indicated by $z_d$ in a differentiable way, a spatial transformer (ST)~\cite{jaderberg2015spatial} is used.
Each skill duration $z_d$ is used starting with the remaining part of the trajectory, i.e., the part of the trajectory that has not been explained by all previous $z_d$.
This extracted sub-trajectory $\bar{\tau}$ is then used as the input for learning the skill library, the generative model parametrized by $\theta$.
Here, we use discrete $\beta$-variational autoencoder (VAE)~\cite{kingma2014auto,higgins2017beta,burgess2018understanding} with skip-connections~\cite{dieng2019avoiding} using the continuous gumbel-softmax/concrete approximation~\cite{jang2016categorical,maddison2016concrete} for the discrete skill type $z_s$, with a latent dimension of size $S$.
This realization of the skill library is very general, which allows to capture any kind of trajectories and is trainable end-to-end within the SKID framework.
Similar VAE based approaches but for single skill trajectories were successfully used in~\cite{osa2020goal,noseworthy2020task}.
The condition of $z^n$ on $z^{n-1}$ is realized as an additional neural layer, with $z^{n-1}$ as input and the output is added to the logits of the encoder network before sampling $z^n$.
The output of the skill library is the sub-trajectory $\hat{\tau}$ approximated with one activated skill, which is then added to the overall approximated trajectory $\tilde{\tau}$ by a spatial transformer.
This iterative process is repeated $N$ times until the sum of $z^n_d$ reaches a given threshold $t_{\epsilon}$, reconstructing a fraction $z_d$ of the full trajectory $\tau$ per step.
The number $N$ is determined by the model through the unsupervised learning process, aiming at reconstructing the data under the constraints of the model.
The full framework is shown in Figure~\ref{fig:model}(b).

\begin{figure*}[t]
\centering
\includegraphics[width=0.90\textwidth]{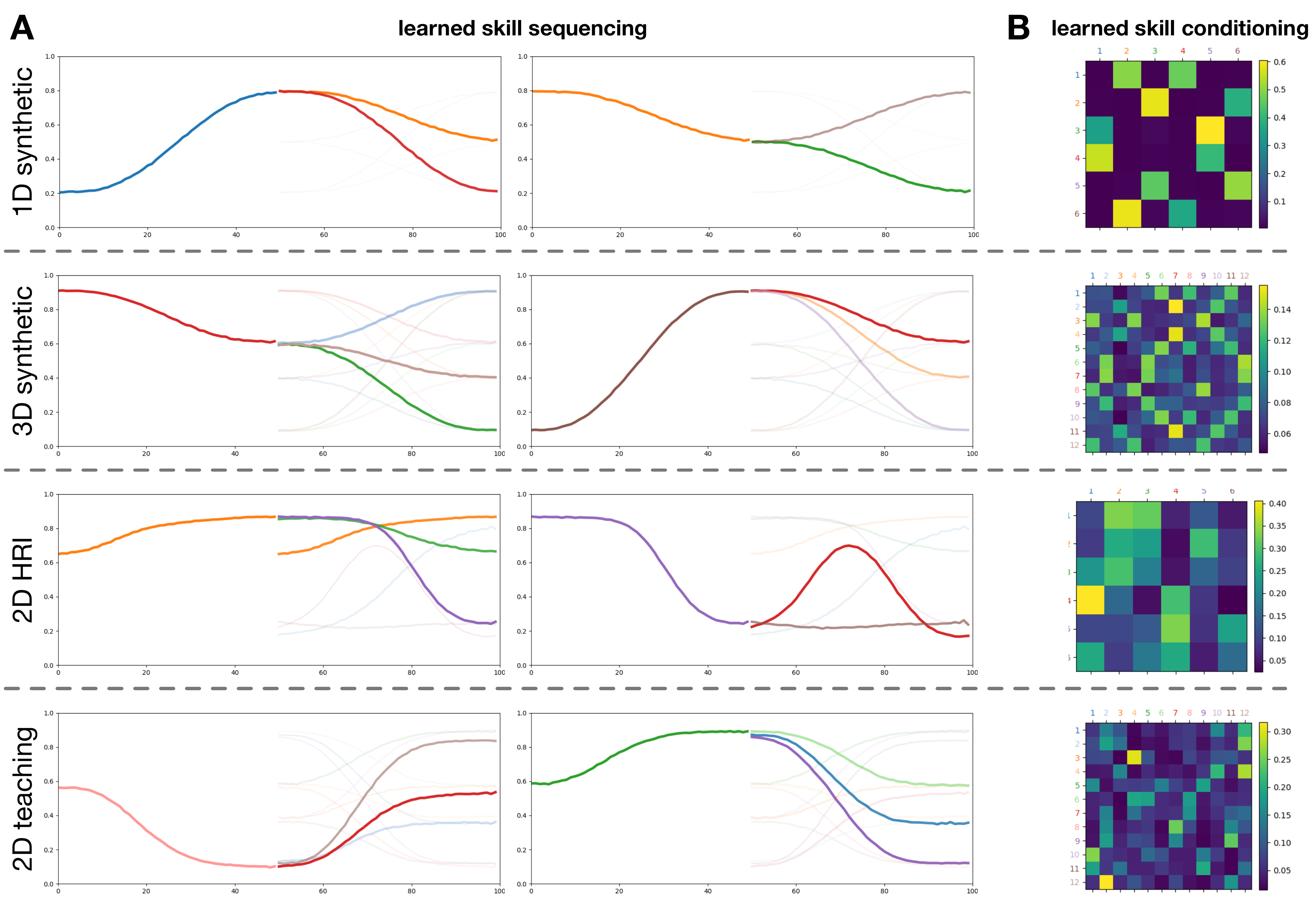}
\caption{Showing the learned skill conditioning by SKID for the four datasets (see Section~\ref{sec:exps}). 
\textbf{A} shows two examples of learned skills and the sequencing of the subsequent skills in one dimension, where the strength of the second plotted skill indicates how likely that skill follows the first one. 
\textbf{B} shows the learned conditioning for all skills as a matrix heatmap, encoded as how likely the skill in column $j$ follows the skill in row $i$.
}
\label{fig:results_condition}
\vspace{-12pt}
\end{figure*}

\subsection{SKID Learning}
Learning is done by jointly optimizing the generative model and the inference network, i.e., the parameters $\theta$ and $\phi$, to maximize the evidence lower bound (ELBO) given as
\begin{align}
\mathcal{L(\tau;\theta,\phi)} &= \mathbb{E}_{q_{\phi}(z | \tau)} \big[ \text{log} \frac{p_{\theta}(\tau , z)}{q_{\phi}(z | \tau)} \big] \nonumber \\
&= \mathbb{E}_{q_{\phi}(z | \tau)} \big[ \text{log} p_{\theta}(\tau | z) \big] - \text{KL}( q_{\phi}(z || \tau) | p(z) ) \ ,
\label{eq:elbo}
\end{align}
with the parametrized likelihood $p_{\theta}(\tau | z)$, the parameterized inference model $q_{\phi}(z | \tau)$ and the latent prior $p(z)$.
The first term aims at reconstructing the data while the KL-divergence forces the model to stay close to a given prior.
Due to the reparametrization trick for the Gaussian distributed $z_d$ and the gumbel-softmax/concrete distributed $z_s$, the $\phi$ parametrized inference network and the $\theta$ parametrized generative model, can be learned jointly via stochastic gradient descent.

To enforce disentangled representations, the $\beta$-VAE~\cite{higgins2017beta} was introduced, which weights the KL term by the hyperparameter $\beta$.
This balancing was further refined by adding capacity terms to the KL~\cite{burgess2018understanding,dupont2018learning}, which can be seen as a slack variable allowing some distance in the KL, and which is increased during training.
Adding these to Equation~\ref{eq:elbo} and separating the different latent variables~\cite{dupont2018learning}, we get our learning objective as
\begin{align}
\mathcal{L(\tau;\theta,\phi)} &= \mathbb{E}_{q_{\phi}(z | \tau)} \big[ \text{log} p_{\theta}(\tau | z) \big] \\
& - \gamma_d | \text{KL}( q_{\phi}(z_d || \tau) | p(z_d) ) - C_d | \nonumber \\
& - \gamma_s | \text{KL}( q_{\phi}(z_s || \tau) | p(z_s) ) - C_s | \nonumber\ ,
\label{eq:elbo_capacities}
\end{align}
with $\gamma_d$, $\gamma_s$ constant scaling factors, and $C_d$, $C_s$ the information capacities, i.e, the allowed slack.

To tighten the lower bound estimate and get a more complex implicit distribution, we use the importance weighted autoencoders (IWAE)~\cite{burda2016importance} objective.
This formulation uses $K$ samples of the latent variables and weights the according gradients by their relative importance.

\begin{figure*}[t]
\centering
\subfigure[1D synthetic.]{\includegraphics[width=.24\textwidth]{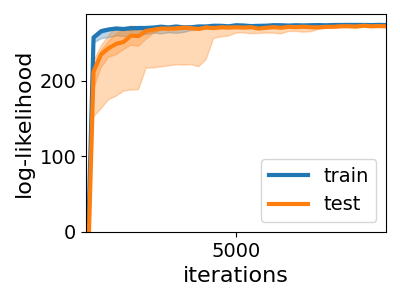}} 
\subfigure[3D synthetic.]{\includegraphics[width=.24\textwidth]{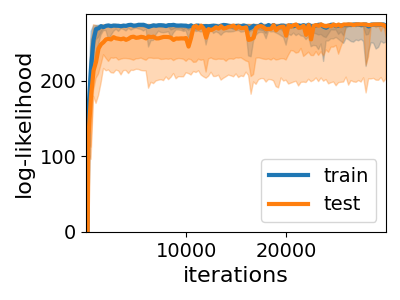}} 
\subfigure[2D HRI.]{\includegraphics[width=.24\textwidth]{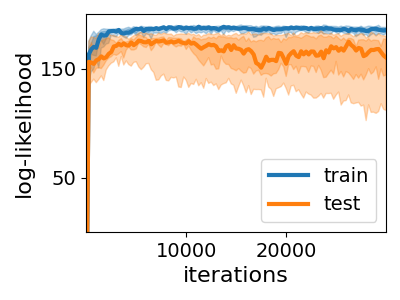}} 
\subfigure[2D teaching.]{\includegraphics[width=.24\textwidth]{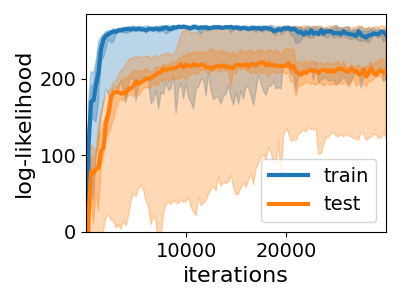}} 
\caption{Shown is the median with the $50\&$ and $90\%$ quantiles of the log-likelihood of the train and test model over $10$ runs.
The test model uses the discrete skill type $z_s$ and is evaluated on a held-out test set, while the train model uses the continuous approximation.
The two real world datasets (bottom row) have a bigger gap between train and test due to the higher noise and variability in the data.
Moreover, the two datasets (right column) with the highest number of required skills (12), tend to get stuck in local optima due to posterior collapse, i.e., no utilizing all discrete dimensions and hence, miss some skills (see also Figure~\ref{fig:missed_skills}(b)).
In addition to these two challenges, the 2D teaching dataset also requires the most complex segmentation of the datasets. 
}
\label{fig:evals}
\vspace{-12pt}
\end{figure*}

\section{Experiments}
\label{sec:exps}
In all experiments, SKID is presented unlabelled trajectories and learns to segment those into skills and the required skills simultaneously.
The different datasets vary in their complexity in multiple ways: artificially created or real world recorded trajectories, the dimensionality of the trajectories, the number of subsequent skills $N$, or the total number of different skills.
Here skills refer to reaching specific locations.
With these different datasets, we test the capacity of SKID to uncover the underlying latent representation from various complex data.
Additionally we test SKID's performance in a continuous learning setting, where the number of skills in the demonstrations is increased during training, and the transfer of skills learned from human demonstrations to the robot.

\paragraph{1D Synthetic}
This dataset consists of 1-dimensional trajectories with up to $N=3$ skills per trajectory, and with a $6$ different skills.
Data was generated by sampling $10.000$ trajectories for each sequence length.
The individual locations per trajectory are sampled uniformly and Gaussian noise is added to each skill location.
The resulting trajectory is generated by creating a minimum jerk trajectory connecting all the locations.
To create a diverse dataset, we adapt a trajectory augmentation method~\cite{osa2020goal} to generate the trajectory $\tau$ as:
\begin{align}
\tau &= \mathcal{N}(\tau_{o}, aB^{\dagger}) \ ,
\end{align}
with the original trajectory $\tau_{o}$, the constant $a$ and $B^{\dagger}$ the Moore-Penrose pseudo-inverse of $M$,  
where $M$ is set to
\begin{align}
M = \begin{bmatrix}
0 & 0 & 0 & \cdots & 0 \\
0 & 2 & -1 & 0 & \vdots \\
0 & -1 & 2 & \ddots & 0 \\
\vdots & 0 & \ddots & \ddots & -1 \\
0 & \cdots & 0 & -1 & 2 \\
\end{bmatrix}
\end{align}
to smoothly propagate the disturbances along the trajectory.
This data augmentation method allows for small datasets and hence less physical demonstrations are required.

\paragraph{3D Synthetic}
This dataset consists of 3-dimensional trajectories with up to $N=3$ skills per trajectory, with a $12$ different skills.
Thus, the dataset is more complex in contrast to the 1-dimensional not only in the dimensionality of the trajectories, but also in the number of skills that need to be learned.
Generating the dataset was done similar to the 1-dimensional dataset.

\paragraph{2D Human-Robot-Interaction}
This dataset is taken from a human-robot-interaction (HRI) scenario presented in~\cite{koert2019learning}, where we use the tracked hand movements of the human to learn the movements used by the human.
As SKID also learns the conditioning or sequencing of the skills, this can be used to predict the human behaviour for a more intelligent robot adaptation in such collaborative scenarios.
The dataset was created by taking the data form one person ($10$ task executions) and using $1000$ random cuts at the four known goal locations, such that each trajectory consists of $2$ to $5$ of those locations, and use the described trajectory augmentation method to create $30.000$ trajectories in total.
The augmented dataset consists of two-dimensional $x,y$ trajectories with up to $N=4$ subsequent skills, and ideally $5$ different skills to learn.
Due to the heuristic cuts and human variation, the actual dataset includes samples with $N > 4$, and has the biggest variability within skills and the biggest noise across the datasets.

\paragraph{2D Teaching}
For this dataset, we recorded the wrist position of a human during an object manipulation task, shown in Figure~\ref{fig:data_record}.
The human was instructed to move between $1$ and $3$ boxes per demonstration between the four locations.
SKID learns the required skills for such manipulation tasks, than can be used for planning by sequencing the discovered skills, and to directly teach the skills to the robot as task space trajectories.
The learned conditioning can be used, for example, to reduce the search complexity of planning algorithms.
In total we recorded $223$ demonstrations and used the trajectory augmentation method described in the synthetic datasets to create a dataset of $5000$ trajectories for each number of moved boxes.
Thus, the resulting dataset consists of two-dimensional $x,y$ trajectories with up to $N=5$ subsequent skills and $12$ different skills to learn.

\begin{table}[!b]
\vspace{-12pt}
\renewcommand{\arraystretch}{0.95}
\caption{Hyperparameters used in the Experiments.}
\label{table:params}
\centering
\begin{tabular}{c||c|c|c|c}
& \bfseries 1D syn. & \bfseries 3D syn. & \bfseries 2D HRI & \bfseries 2D teaching \\
\hline\hline
\bfseries RNN size & 64 & 64 & 64 & 32 \\ 
\bfseries linear size & 16 & 16 & 16 & 16 \\
\bfseries $N_{\text{max}}$ & 3 & 3 & 4 & 5 \\
\bfseries $\alpha$ & $1e^{-3}$ & $1e^{-3}$ & $1e^{-3}$ & $5e^{-4}$ \\
\bfseries $\lambda$ & $5e^{-2}$ & $5e^{-1}$ & $5e^{-3}$ & $5e^{-1}$ \\
\bfseries $C_d$ 	& $(0,1,30k)$ & $(0,2,30k)$ & $(0,2,30k)$ & $(0,2,30k)$\\
\bfseries $C_s$ 	& $(0,1,30k)$ & $(0,1,30k)$ & $(0,1,30k)$ & $(0,1,30k)$\\
\bfseries $\gamma_d$ & $5$ & $3$ & $10$ & $3$\\
\bfseries $\gamma_s$ & $5$ & $3$ & $10$ & $3$\\
\bfseries $\omega$ & $(1,0.2,30k)$ & $(1,0.2,30k)$ & $(1,0.2,30k)$ & $(2,0.5,30k)$\\
\bfseries $\mu_{d}$ & $0$ & $0$ & $0.5$ & $-0.5$\\
\bfseries $\sigma_{d}$ & $1$ & $1$ & $1$ & $1$\\
\bfseries \texttt{VAE} $\sigma$ & $0.02$ & $0.02$ & $0.02$ & $0.02$\\
\bfseries $S$ & $6$ & $12$ & $6$ & $12$\\
\bfseries $p_{\theta}$ $\sigma$ & $0.1$ & $0.1$ & $0.15$ & $0.1$\\
\hline
\end{tabular}
\end{table}

\subsection{Experimental Setup}
Learning is done by optimizing Equation~\ref{eq:elbo_capacities} with the Adam optimizer~\cite{kingma2015adam} with learning rate $\alpha$, weight decay $\lambda$~\cite{loshchilov2019decoupled}, with batch prior regularization~\cite{zhao2018unsupervised} for the discrete skill type, and the loss is variance normalized.
The capacities $C_d$, $C_s$ are linearly increased during training from (start,end,iterations).
Parameters were optimized with grid search for the $1$D synthetic data set, and modified for the other datasets with smaller grid searches.
The mini-batch size is set to $64$ and for the IWAE importance sampling $k = 20$ samples are used.
The recurrent neural network (RNN) that learns $z_d$ is a vanilla recurrent network, the threshold is set to $t_{\epsilon} = 0.85$ and maximum iterations is set to $N_{\text{max}}$.
The skill library VAE encoder and decoder consist of two hidden layers with $[30,15]$ neurons with \texttt{elu} activations, latent layer of size $S$ (skill library size), and skip connections within the decoder.
For the gumbel-softmax encoded $z_s$ the temperature $\omega$ is decreased from (start,end,iterations)
with a cosine schedule, using the continuous approximation for training but hard one-hot vectors for evaluation.
Trajectories are normalized to $200$ steps and the sub trajectories extracted by the spatial transformer for the skill library consist of $50$ steps.
The remaining parameters for all datasets are given in Table~\ref{table:params}.

\begin{figure}[t]
\centering
\hfill
\subfigure[Maximum log-likelihood.]{\includegraphics[width=.23\textwidth]{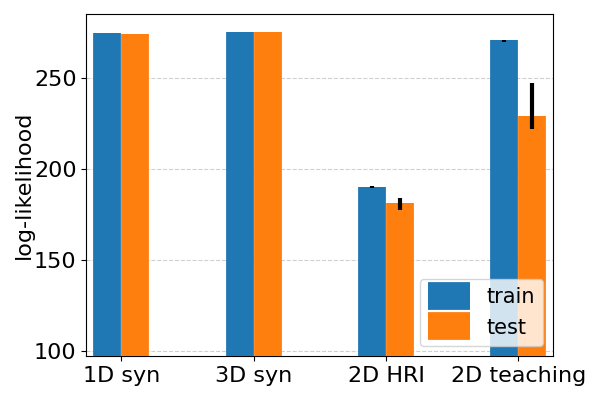}} 
\hspace{0pt}
\subfigure[Unidentified skills.]{\includegraphics[width=.23\textwidth]{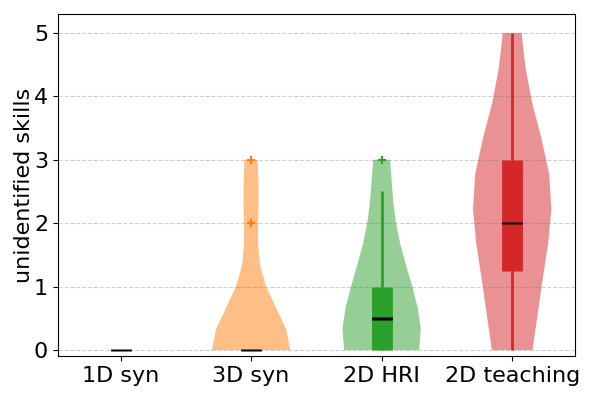}}
\hfill
\vspace{-2pt}
\caption{(a) Shown is the median and interquartile range of the maximum log-likelihood of the train and test model over $10$ runs for each dataset.
The test model uses discrete one-hot encodings for $z_s$ and is evaluated on a held-out test set.
(b) Shown are the number of missed skills, i.e., skills that were not recognized and learned, for each dataset over $10$ runs each.
Black horizontal line indicates the mean, the box the interquartile range with whiskers, crosses show outliers, and the shaded area represents the data density.
Identifying skills is harder in the real world datasets due to their higher noise and variability.
In addition, the 2D teaching dataset also has the most complex segmentation and like the 3D synthetic dataset, the most required skills ($12$).
}
\label{fig:missed_skills}
\vspace{-12pt}
\end{figure}

\subsection{Results}
The goal of SKID is to simultaneously learn to segment trajectories into a varying number of sub-trajectories (skills) and these skills from unlabelled trajectories.
Additionally, as part of the skill library, a skill conditioning is learned, that encodes how likely a certain skill follows after another skill.
In Figures~\ref{fig:results_overview}~\&~\ref{fig:results_condition} qualitative results on the different datasets are summarized, showing the evaluation of the test model on held-out data and using hard one-hot samples of $z_s$.
Evaluations over $10$ runs per dataset are shown in Figure~\ref{fig:evals}~\&~\ref{fig:missed_skills}, where the log-likelihood and number of unidentified skills are shown.

\begin{figure*}[t]
\centering
\includegraphics[width=0.95\textwidth]{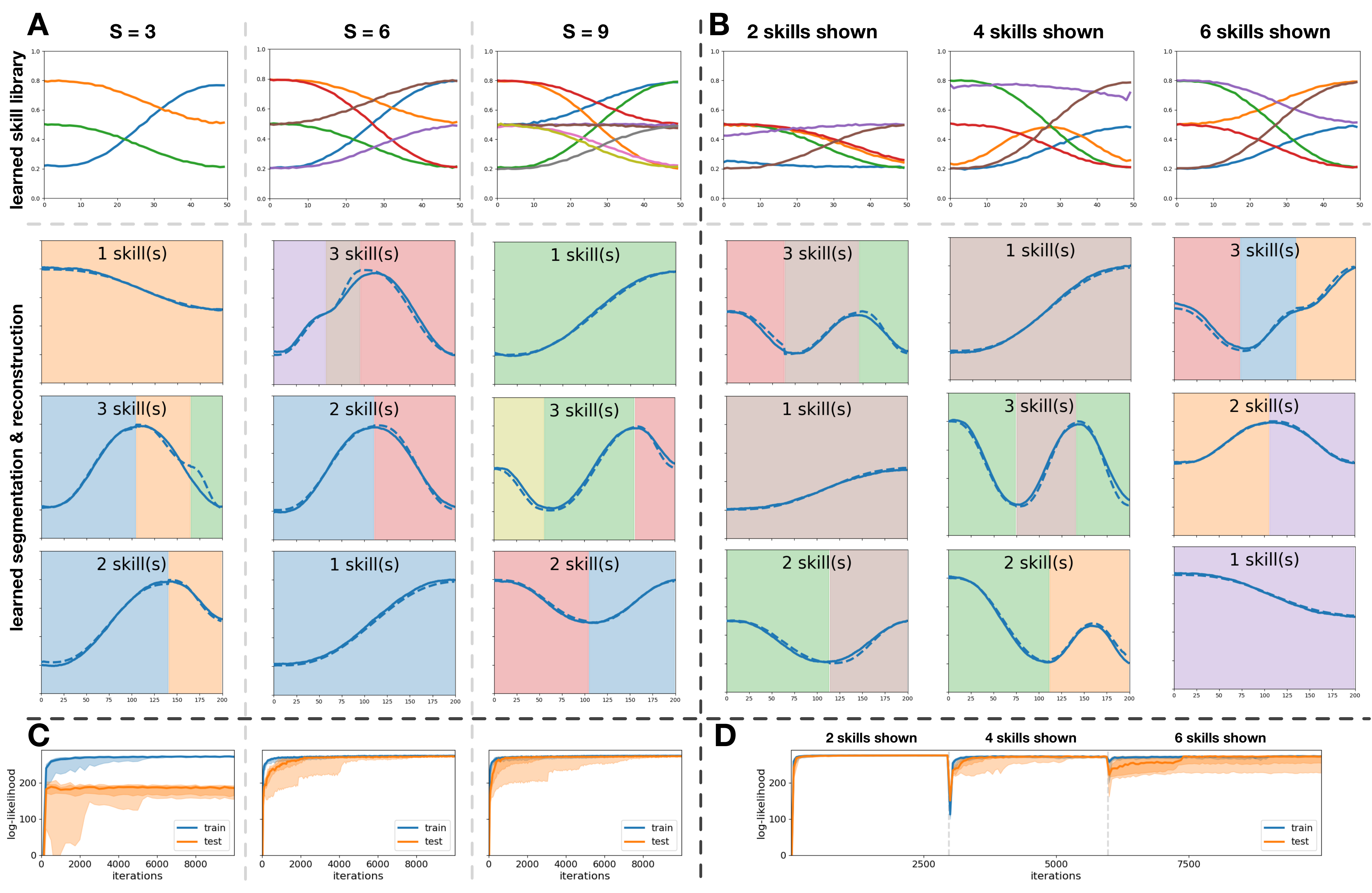}
\caption{Showing results on the $1$D synthetic dataset with different latent dimension sizes $S$, i.e., skill library size, (\textbf{A}) and when increasing the shown skills in the demonstrations during training (\textbf{B}).  
Each column shows one setting for $S$ or the number of currently shown skills, top row shows the learned skill library, and rows below show sampled demonstrations with their reconstruction (dashed) and segmentation (colored background indicate the corresponding used skill from the library above).
\textbf{C-D} shows the median with the $50\&$ and $90\%$ quantiles of the log-likelihood of the train and test model over $15$ runs.
The test model uses the discrete skill type $z_s$ and is evaluated on a held-out test set, while the train model uses the continuous approximation.
Dashed lines in \textbf{D} indicate when more skills were added to the demonstrations.
}
\label{fig:additional_evals}
\vspace{-10pt}
\end{figure*}

\subsubsection{Segmentation \& Skill Library}
The segmentation and skill library learning results for all four datasets are shown in Figure~\ref{fig:results_overview}, where each row corresponds to one dataset.
The plot shows one exemplary run for each dataset.
In the left part of the figure (A), different random trajectories are shown along with their reconstruction and segmentation.
Line colours indicate the different dimensions of the trajectories, where solid lines show the input trajectory $\tau$ and dashed lines show the reconstructed trajectory $\tilde{\tau}$.
Shaded areas indicate the skill used for that segment, where the according learned skills are shown in the right part of the figure (B), with one panel per trajectory dimensions and one coloured line per skill.

For all datasets SKID is able to learn to segment given trajectories into skills and the required skill library simultaneously without supervision.
Learning the underlying skills and their segmentation from full plan executions without additional knowledge and feedback, enables users to teach a robot sequential tasks in a natural way by just executing the full task, without having to specify and demonstrate the individual skills. 
This learned knowledge can then be used by the robot to, for example, solve new task instantiations by planning with the learned skills.

\subsubsection{Skill Conditioning}
In addition to learning the segmentation and skill library, SKID also learns a skill conditioning.
This learned conditioning and sequencing is shown for all datasets in Figure~\ref{fig:results_condition}, with each dataset in one row.
In the left part (A), two examples are shown in the two panels.
In each panel, first one learned skill is plotted followed by all learned skills, but with the line intensity scaled by how likely this skill follows the first one.
This sequencing shows that SKID is able to learn which skills are likely to follow each other, and that, for example, jumps between subsequent skills are unlikely.
The right part of the figure (B) shows this learned skill conditioning for the whole skill library with a matrix heatmap, where the values indicate how likely the skill in column $j$ follows the skill in row $i$.

SKID is able to learn this skill conditioning, i.e., task structures, for all datasets.
This information can additionally be used, for example, to support the planning for new tasks with the learned skills by reducing the planning space.
Another possibility for human-robot-interaction scenarios, the robot can learn to \textit{understand} the movements of the human, and use the learned structure to predict the future behaviour of the human in order to adapt its own behaviour.

\subsubsection{Continuous Learning}
In many settings not all required skills are known when starting to teach a new task.
Hence, the number of skills shown in the demonstrations may increase over time and the train data distribution changes.
We tested SKID's ability to learn continuously in such a setting, when the number of skills in the demonstrations is increased during training.
The results are shown in Figure~\ref{fig:additional_evals}, using the $1$D synthetic dataset and increasing the number of skills in the demonstrations every $3000$ iterations from $2$ to $4$ to $6$.
The model successfully learns all skills in the corresponding phase and is able to adapt the skill library when new skills are presented.
Note, the same parameters and settings as in the previous experiments were used, except the temperature decay is reset when the data distribution changes.
By changing the hyperparameters for such settings and incorporating additional mechanisms from lifelong-learning research, the performance and flexibility of SKID in continuous learning scenarios may even be increased.

\subsubsection{Robot Skill Execution}
\begin{figure*}[t]
\centering
\includegraphics[width=0.97\textwidth]{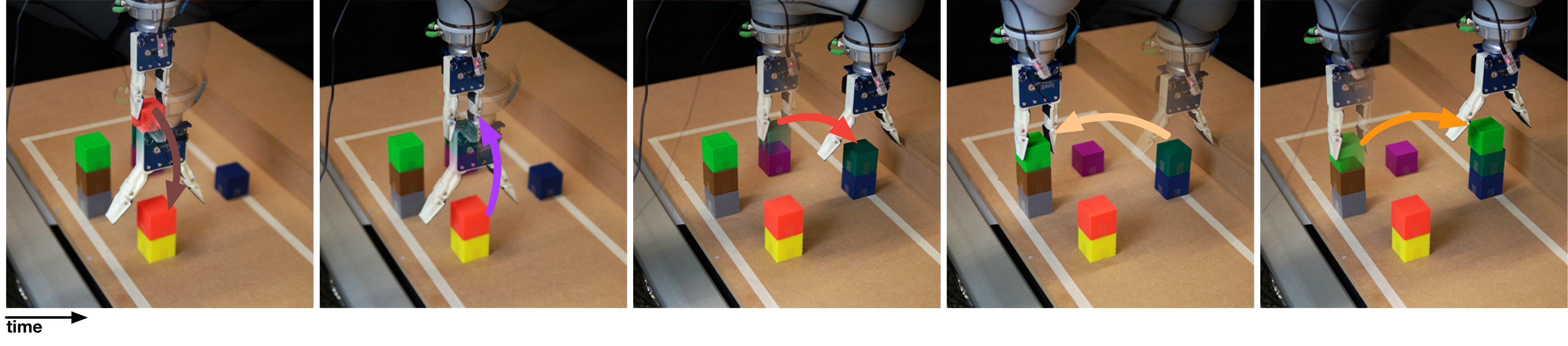}
\caption{Snapshots of the robot execution of a sequence of skills learned from human demonstrations ($2$D teaching) using SKID.
Coloured arrows indicate the used skill from the learned library shown in Figure~\ref{fig:results_overview}B.
The learned skills represent the xy-transition trajectories between the four stacking positions.
}
\label{fig:robot_seq}
\vspace{-12pt}
\end{figure*}

To test the transfer of the learned skill library, we used the skills learned from the human $2$D teaching demonstrations to rearrange objects with a real KUKA robot with a SAKE gripper.
Snapshots of the robot execution of a sequence of learned skills are shown in Figure~\ref{fig:robot_seq}.
The learned skills -- as shown in the skill library in Figure~\ref{fig:results_overview}B last row -- represent the xy-transition trajectories between the four stacking locations.
Each skill duration is set to $5$ seconds for execution, height adjustment and gripper control is automated.
The skills learned autonomously from unlabelled raw human trajectories can be successfully used on the real robot.

\subsection{Limitations}
While SKID is able to learn the segmentation, conditioning and skill library for different datasets with varying complexities, the stochastic variational inference setting is challenging.
Not all runs achieve perfect results (see Figures~\ref{fig:evals} \& \ref{fig:missed_skills}), where the major problem that occurs is that the discrete VAE used for the skill library sometimes \textit{misses} one or a few skills (see Figure~\ref{fig:missed_skills}(b)), even with good or perfect segmentation.
In other cases the unidentified skills hinder the learning of perfect segmentation (see Figure~\ref{fig:missed_skills}(a)).
Due to the continuous approximation during training, the training model can mix multiple skills to achieve good performance, the test model with hard one-hot vectors for the skills type $z_s$ can only use one skill, and thus, this results in a sub-perfect performance.
While using lower temperatures for the gumbel-softmax during training can help with this issue, the lower temperatures create higher variance in the gradients.

As the number of skills has to be specified for the skill library, another possibility is to use more skills than required (higher $S$), such that unused latent dimensions matter less.
This strategy was successful in some experiments (especially in the 2D HRI setting and see Figure~\ref{fig:additional_evals} for an evaluation of different settings of $S$), but with more potential skills, the segmentation may become worse, as the model tends to learn too complex skills, i.e., combining multiple skills into one, as the higher number of available skills reduces the necessity of finding the simplest individual skills, or learning \textit{uninformative} skills used as connection skills, e.g., holding a position (see Figure~\ref{fig:additional_evals}), and hence may lead to over-segmentation. 
The size of the skill library influences the quality of the segmentation.

While the choice of using a VAE for the skill library is very general and allows straightforward learning of any trajectories, it has the disadvantage of a fixed latent space, i.e., number of (maximum) skills, and less flexible skills.
The SKID framework is not bound to this choice, and more complex realizations like goal-conditioned VAE's~\cite{osa2020goal} or a set of probabilistic movement  primitives~\cite{paraschos2018using} can be incorporated if more flexible skills are required.

\section{Conclusion}
We proposed a novel Bayesian approach to simultaneously learn trajectory segmentation and skills from unlabelled raw trajectories.
The SKID framework builds on a hierarchical VAE structure and learns simultaneously to segment trajectories into reoccurring patterns, the skills to reproduce them, and the temporal relation between these skills.
These features were successfully shown on multiple datasets with varying complexities, including two datasets with tracked human motions, and transferred to a robotic setup.

Such automatic skill discovery can be used as a natural interface for teaching a robot complex tasks consisting of sequences of multiple skills. 
In addition with the learned skill conditioning, the framework can also be used to analyse and predict the behaviour of a human, or another robot, for a more intelligent adaptive behaviour of the agent.

The next challenge is to increase the robustness of the learning, especially in settings with a high number of skills, continuous learning settings, and noisy real world data.
Moreover, the framework is not limited to robotic or human movement trajectories, but can be applied to any kind of trajectories, that consist of reoccurring patterns and opens interesting future research.
Another promising direction could tackle the limitation of the offline setting, i.e., full trajectories are used for segmentation and learning, by integrating ideas from online change point detection approaches~\cite{adams2007bayesian,agudelo2020bayesian} into the SKID framework and feeding the trajectories step by step to allow the processing of longer sequences.




%
%


\bibliographystyle{ieeetr}
\bibliography{skid_bib}

\end{document}